%% file: tmlr.tex
\documentclass[10pt]{article} 
\usepackage[preprint]{tmlr}

\input{math_commands.tex}

\usepackage{url}
\usepackage[pagebackref,breaklinks,colorlinks]{hyperref}
\usepackage{graphicx}
\usepackage{booktabs}
\usepackage{makecell}

\title{When LLaVA Meets Objects: Token Composition for Vision-Language-Models}

\author{\name Soumya Jahagirdar \email soumya-shamarao.jahagirdar@uni-tuebingen.de \\
      \addr Tuebingen AI Center/University of Tuebingen
      \AND
      \name Walid Bousselham \email walid.bousselham@gmail.com \\
      \addr Tuebingen AI Center/University of Tuebingen
      \AND
      \name Anna Kukleva \email akukleva@mpi-inf.mpg.de \\
      \addr Max Planck Institute for Informatics, SIC
      \AND
      \name Hilde Kuehne \email h.kuehne@uni-tuebingen.de\\
      \addr Tuebingen AI Center/University of Tuebingen \\
      MIT-IBM Watson AI Lab}

\def\month{MM}  
\def\year{YYYY} 
\def\openreview{\url{https://openreview.net/forum?id=XXXX}} 

\begin{document}

\maketitle

\begin{abstract}
Current autoregressive Vision Language Models (VLMs) usually rely on a large number of visual tokens to represent images, resulting in a need for more compute especially at inference time. 
To address this problem, we propose Mask-LLaVA, a framework that leverages different levels of visual features to create a compact yet information-rich visual representation for 
 autoregressive VLMs. 
Namely, we combine mask-based object representations together with global tokens and local patch tokens. While all tokens are used during training, it shows that the resulting model can flexibly drop especially the number of mask-based object-tokens at test time, allowing to adapt the number of tokens during inference without the need to retrain the model and without a significant drop in performance. 

We evaluate the proposed approach on a suite of standard benchmarks showing results competitive to current token efficient methods and comparable to the original LLaVA baseline using only a fraction of visual tokens. 
Our analysis demonstrates that combining multi-level features enables efficient learning with fewer tokens while allowing dynamic token selection at test time for good performance.\footnote{All code and evaluation scripts will be made publicly available.}
\end{abstract}

\section{Introduction}

Autoregressive Vision-Language Models (VLMs) have recently become one of the most popular model families for a broad variety of vision-language tasks, ranging from image captioning \cite{show_and_tell_cvpr_2015,flicker_dataset,conceptual_captions_acl_2018} to Visual-Question-Answering (VQA) \cite{vqav2,vizwiz,gqa}.
These models usually consist of a visual encoder~\cite{learning_transferable_vis_icml_2021} that converts an input image to serialized output along with LLM that generates text response. 
To enable the LLM to process visual inputs, an alignment module bridges the gap between visual and textual representations, typically implemented as a simple linear layer~\cite{visual_intruction_tuning_2023}, an MLP projector~\cite{Improved_baselines_with_visual_instruction_tuning_cvpr_2024}, or a query-based network~\cite{qwen_vl,blip_2_icml_2023}.
 While those models reach impressive performance across a huge variety of benchmarks, they usually rely on fine-grained image representations to achieve this is performance.
Namely, LLaVA-1.5~\cite{visual_intruction_tuning_2023} encodes an image with {$336\times336$} resolution into 576 visual tokens, while LLaVA-NeXT~\cite{llava_next_2024} and mini-Gemini-HD~\cite{minigemini} double the resolution, resulting in 2880 tokens resulting in significantly more visual tokens than the text prompt tokens. 
Consequently, the number of tokens, and thus the inference cost on the language models side increases significantly.

Several works tried to address this problem e.g. by focusing on token pruning~\cite{fastv_eccv_2024,fitprune_2024,sparsevlm_2024,faster_vlm_2024}, thus dropping tokens from the visual stream, or compression and pooling~\cite{funnel_transformer_nips_2020,pyramid_bert_2022,dynamic_token_pooling_2022,compressive_transformer_2019}, thus combining the representation of some or all tokens into fewer ones. Usually, those methods operate on the patch-wise tokens of the underlying vision backbone. 
To our best knowledge, no prior work has attempted to combine CLS or patch tokens with object-centric representations in on input stream so far.

In this context, we propose Mask-LLaVA, a VLM that combines representations of different granularity into an efficient yet meaningful input for state-of-the-art LLMs.
Namely, we feature three different types of inputs: CLS tokens, pooled patch level tokens, and mask-based object tokens from on automatically generated object masks.
While the CLS token is the direct output of the vision backbone, and the pooled patch level tokens are simply based on an average pooling operation over the original patch tokens from the vision backbone, we propose an automated pipeline to gather object specific representation. To this end, we first run an automated objectness detection~\cite{deformable_detr_2020}, keeping 100 bounding box proposals with the highest confidence, followed by a segmentation~\cite{sam_iccv_2023}, and a region representation learning~\cite{maskinversion_2024} step.  Practically, the detected and segmented bounding boxes show a certain redundancy with some masks overlapping by a significant fraction, thus technically oversampling the overall object count of the image. 
We train the respective model on the concatenated input of all tokens, resulting in 75\% fewer tokens compared to the LLaVA baseline. \newline

Our experiments demonstrate that once trained on this oversampled content, the model allows for flexible reduction of object tokens while maintaining—and in some cases even improving—performance.

Analysis of token characteristics reveals that different token types exhibit varying norms: patch tokens possess the highest values, mask-based object tokens intermediate values, and CLS tokens the lowest.
We therefore propose a token scaling to normalize the token norm of CLS and object tokens to the level of patch tokens. Our experiments indicate that normalizing the norm of tokens per image leads to the best overall results in this case.

We evaluate the proposed framework on a suite of eight benchmark datasets, VQAv2 \cite{vqav2}, GQA \cite{gqa}, VizWiz \cite{vizwiz}, ScienceQA-IMG \cite{scienceqa}, POPE \cite{pope}, MME \cite{mme}, MMBench \cite{mmbench}, and MM-Vet \cite{mmvet}. Results show that the proposed representation is highly competitive compared to other approaches in the field. Our ablation further evaluates the scaling, as well as the token composition and the impact of pretraining on the full token set. 
While patch tokens alone establish a strong baseline, our experiments confirm that the additional design elements further enhance performance.

We summarize the contributions of this work as follows: \\
(1) We propose Mask-LLaVA as an efficient VLM framework which allows varying number of tokens at test time.\\
(2) To this end, we leverage a combination of global CLS and local patch tokens together with mask based object specific tokens and propose a norm scaling to allow the different groups of tokens to be processed jointly.\\
(3) We provide an extensive evaluation on various challenging benchmarks showing how training with oversampling of masked-based object representations can increase the flexibility with respect to token counts.\\

\section{Related Work}
\label{related_works}

\subsection{Multimodal Vision Language Models}

The success and applications of large language models (LLMs) 
\cite{language_models_are_few_shot_learners_nips_2020, llama_3_2024, dataset_Analyst_Subpopulation_Structure_eccv_2024,Training_language_models_to_follow_instructions_nips_2022,llama_2023,vicuna_2023,Language_models_are_unsupervised_multitask_learners} have resulted in the growth of extending their reasoning capabilities to multi-modal tasks, leading to the emergence of vision-language models (VLMs)\cite{language_is_not_all_you_need_2023, Palme_2023, visual_intruction_tuning_2023, qwen_vl, blip_2_icml_2023, li2022blip}. 
BLIP \cite{li2022blip} uses ViT \cite{visual_intruction_tuning_2023} and divides an image into patches and encodes them as sequence of embedding with [CLS] token. It proposes Image-grounded text encoder, which injects visual information by inserting and an additional cross-attention layer between the self-attention layer. 
BLIP-2 \cite{blip_2_icml_2023} and InstructBLIP \cite{instruct_blip_nips_2023} use a QueryFormer (Q-Former) as the trainable module to bridge the gap between the frozen image encoder and the frozen LLM. Another line of work, like LLaVA \cite{visual_intruction_tuning_2023}, introduces multimodal projector (MLP) to align visual features with text features.

\subsection{Token Pruning for VLMs}
In order to have faster inference for VLMs, visual tokens which occupy the majority of the input sequence can be optimized. Multiple token pruning methods such as~\cite{fastv_eccv_2024,fitprune_2024,sparsevlm_2024,faster_vlm_2024} have been proposed to improve the model inference efficiency by removing less important tokens from the sequence. FastV~\cite{fastv_eccv_2024} identifies the redundancy in visual tokens by removing visual tokens with low attention scores after layer 2 of LLM. FitPrune~\cite{fitprune_2024} introduces a method to fit pruning recipes based on attention statistics. SparseVLM~\cite{sparsevlm_2024} removes distractions from text prompts and uses more accurate text attention to sparsify visual tokens progressively. FasterVLM~\cite{faster_vlm_2024} proposes a method where it shows that text-visual attention does not align well with visual token importance and instead use only [CLS] attention from the visual encoder which yields better results. 

\subsection{Token Compression for VLMs}

Multiple studies have explored token sequence compression in language models~\cite{funnel_transformer_nips_2020,pyramid_bert_2022,dynamic_token_pooling_2022,compressive_transformer_2019}. 
Compared to text image information tends to have higher redundancy, making visual token compression more reasonable and effective in VLMs. LLaVA-PruMerge~\cite{llava_prumerge_2024} uses attention mechanism to select important visual tokens and merges them using similar key clustering, and achieves competitive performance while improving VLM inference efficiency. LlaVolta~\cite{Llavolta_2024} proposes a heuristic and stage-wise compression method that reduces VLM training costs while maintaining original performance. Llama-vid~\cite{llama_vid_eccv_2024} uses context tokens in Q-Former or applies adaptive pooling at the patch level. LLaVA-Mini~\cite{llava_mini_2025} proposes a modality pre-fusion which fuses visual information into text tokens in advance, which consists of a few trainable transformer layers.

\section{Mask-LLaVA}
\label{maskllava}

In this section, we introduce Mask-LLaVA, an efficient large multimodal model that processes visual inputs at three distinct granularities. Similar to previous works, Mask-LLaVA employs a vision encoder to transform images into tokens that capture various levels of information. Specifically, we include:
\begin{enumerate}
    \item The \textbf{[CLS]} token as a \textbf{global context} representation.
    \item \textbf{Pooled patch tokens} to encode \textbf{local context}.
    \item \textbf{Mask-based object tokens} to represent \textbf{object-level} features.
\end{enumerate}

Tokens from these three levels are passed into a multimodal projector, which enables the fusion of features. By incorporating hierarchical representations, Mask-LLaVA significantly reduces the number of tokens required while maintaining performance comparable to LLaVA which processes all patch tokens from ViT, thereby eliminating redundancy and better feature representations.  

\begin{figure*}[t]
  \centering
   \includegraphics[width=1.0\linewidth]{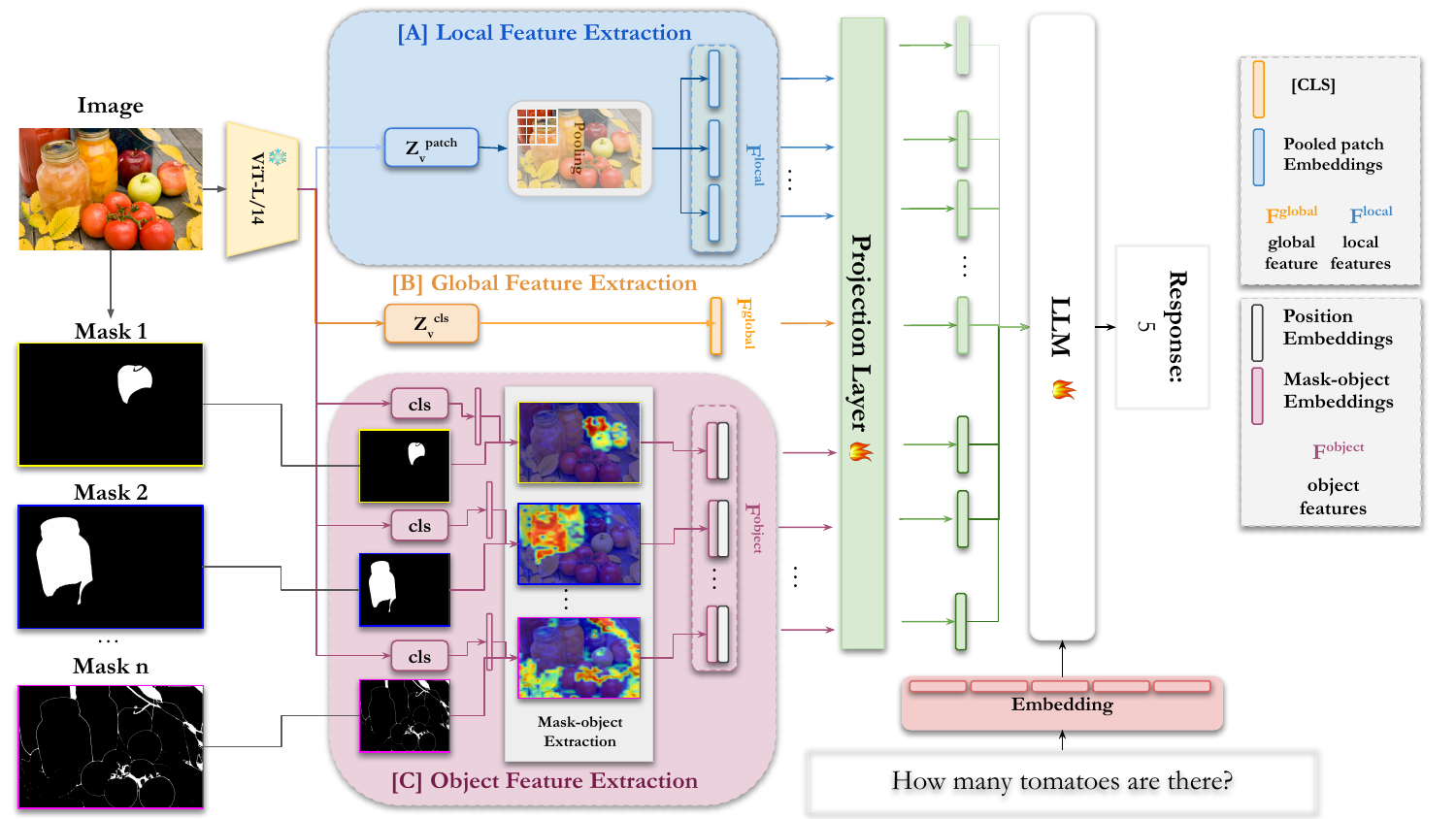}

   \caption{\textbf{Overview of Mask-LLaVA Architecture.} Given an input image, the local feature extraction module pools patch tokens from the Vision Transformer ViT~\cite{learning_transferable_vis_icml_2021} based on 2D grid structure to obtain local context features. Simultaneously, the SAM~\cite{sam_iccv_2023} generates masks, which are used in the object feature extraction module along with the [CLS] token from ViT to obtain mask-based object representations. The explainability map from the [CLS] token is guided to focus on the corresponding masked regions. Finally, the [CLS] token, pooled local features, and object features are projected and passed to LLM along with a question to generate a response.}
   \label{fig:maskllava_arch}
\end{figure*}

\subsection{Visual Token Composition}
An overview of Mask-LLaVA is shown in Fig. \ref{fig:maskllava_arch}. Given an image $I$, we follow the approach of LLaVA and utilize a pretrained visual encoder to extract visual features as: 
\begin{equation}
    Z_v = g(I),  Z_v  = (Z_v^{\text{cls}}, Z_v^{\text{patch}}),
\end{equation}
where \( Z_v \) represents the visual tokens, with the CLS token $Z_v^{\text{cls}}$ and the patch tokens $Z_v^{\text{patch}}$, extracted from the image using a pretrained vision transformer model.
Based on that, Mask-LLaVA computes features at three granularities:

\begin{enumerate}
    \item Global Feature Vector $(F_{global})$: The $[CLS]$ token of ViT serves as representation that contains the global context of the image with $F_{\text{global}} = Z_v^{\text{cls}}$. 
    \item Local Feature Vectors $(F_{local})$: The 576 patch tokens produced by ViT are first reshaped into a $24 \times 24$ 2D grid, preserving their spatial structure. To reduce the redundancy and obtain a more compact local representation, we apply 2D average pooling, $F_{\text{local}} = \mathcal{AP}(Z_v^{\text{patch}})$ with a kernel size of 4, resulting in pooled patch tokens that effectively capture local information. 
    \item  Object Feature Vectors $(F_{object})$: Object-level features are extracted by processing different segmentation masks in the image. For each segmentation mask, a feature vector is generated such that it can produce an explainability map focusing on the corresponding region, as explained in \ref{subsection_mask_token_computation}.
\end{enumerate}

\subsection{Mask Token Computation}
\label{subsection_mask_token_computation}
\begin{figure}[t]
  \centering
   \includegraphics[width=0.5\linewidth]{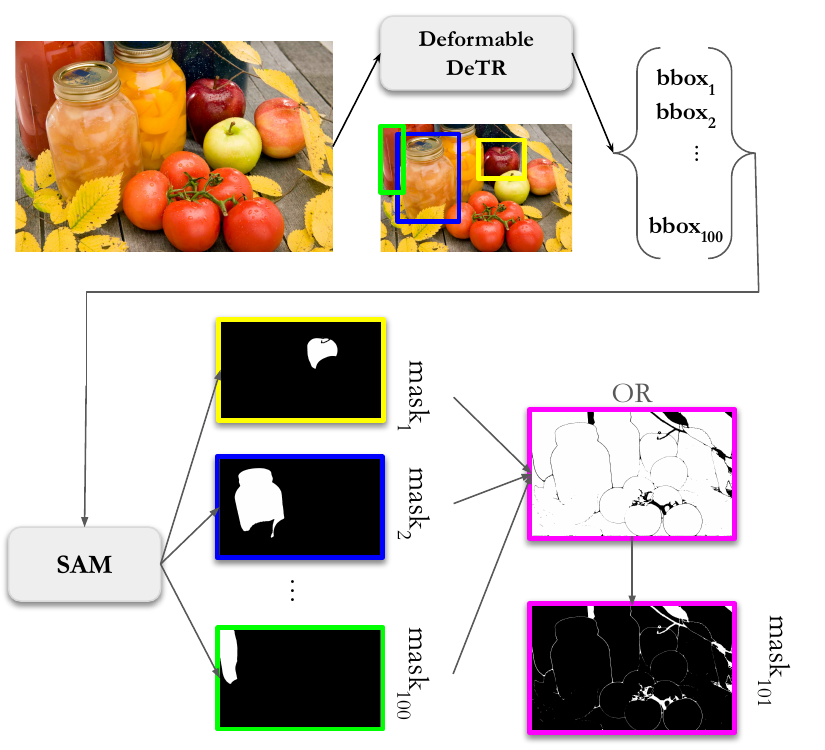}

   \caption{\textbf{Mask-Token Computation.} This figure illustrates the process of obtaining segmentation masks. First, an objectness detector~\cite{deformable_detr_2020} identifies bounding boxes in the image. These bounding boxes, along with the image, are then passed to the SAM~\cite{sam_iccv_2023} model to generate segmentation masks. Additionally, a background mask is included. The resulting masks are then used to extract mask-based object features.}
   \label{fig:object_feat_module}
\end{figure}

To compute the respective mask-based token representation $F_{object}$ for a given image $I$, we first employ an objectness detector~\cite{deformable_detr_2020} to obtain the bounding boxes for the objects present in the scene. These bounding boxes, along with their respective images, are then fed to SAM~\cite{sam_iccv_2023} to generate segmentation masks. This process is illustrated in Fig. \ref{fig:object_feat_module}. Additionally, based on all resulting masks, we add one more a background mask to cover all regions of the image not included in any of SAM's segmentation outputs, ensuring complete coverage of the image and leading to $M = \{m_1, ..., m_n\}$ final segmentation masks. An overview of the pipeline is shown in Figure~\ref{fig:object_feat_module}.

To obtain object embeddings for the respective masks, we leverage Maskinversion~\cite{maskinversion_2024}. Specifically, Maskinversion uses a query mask alongside the original image and $[CLS]$ token from ViT~\cite{learning_transferable_vis_icml_2021} to learn a local representation corresponding to the query mask. This is done by optimizing the respective token so that the explainability map of the token corresponds to the segmentation mask by iteratively updating the embedding token. The update process thus minimizes the discrepancy between the explainability map and the query mask, thereby aligning the embedding representation with the segmented region. To incorporate spatial information into object tokens, we use a fixed sinusoidal positional embedding based on the center coordinates of bounding boxes. This ensures that the model captures relative positional relationships effectively. Denoting the object representation function as $g(\cdot)$, a single token $F_{\text{object}_m}$ for mask $m$ is generated as:

    \begin{equation}
        F_{\text{object}_m} = g(m, I, Z_v)
    \end{equation}

We perform this for all segmentation masks generated by SAM as well as for the background mask \( M = \{m_1, ..., m_n\} \). Note that while we employ a single mask notation here for compatibility, Maskinversion supports gradient decomposition, where all masks for a single image can be computed based on one explainability map, resulting in the joint computation of all masks for one image. Originally designed for localized embeddings of specific image regions, we repurpose Maskinversion to decompose a visual backbone's feature representation into its constituent object parts, introducing its novel use in multimodal learning.



\subsection{Scaling}
To ensure the consistency in scale across the different token types, we normalize the CLS as well as the object tokens on the mean and standard deviation of the norms of the patch tokens. We experiment with different types of scaling, which are explained in detail in the Evaluation section \ref{sec:evaluation}. The global feature is scaled as follows: $\hat{F}_{global} = {F_{global}} \cdot \sigma_{local} + \mu_{local}$, and the object features are scaled as follows: $\hat{F}_{object} = {F_{object}} \cdot \sigma_{local} + \mu_{local}$
where $\mu_{local}$ and $\sigma_{local}$ are the mean and standard deviation of the patch tokens.

\subsection{Architecture and Training}
Following the architecture of LLaVA, Mask-LLaVA incorporates a projector $\mathcal{P}(\cdot)$  designed to align visual representations (in our case, global, local and object features) with the text space. The output of this projector, along with the text embeddings, is then fed into the LLM, which processes these inputs to generate text.
\begin{align}
    H_v &= \mathcal{P}\Big(\hat{F}_{\text{global}}, \ F_{\text{local}}, \ \hat{F}_{\text{object}} \Big)
\end{align}
\begin{align}
    Y &= \mathcal{LLM}\big( H_v, H_t \big),
\end{align}
where $H_t$ is the text embedding of the language prompt.

\paragraph{Training:}

Mask-LLaVA follows the same training process of LLaVA, which consists of two stages:

\begin{itemize}
    \item \textbf{Stage 1: Vision-Language Pretraining} We first train the projector $\mathcal{P}(\cdot)$ to align vision representations with different granularities and language representations using captioning data. The training only focuses on the projection module while the vision encoder and LLM are kept frozen. 
    \item \textbf{Stage 2: Instruction Tuning} Second, the projection layer $\mathcal{P}(\cdot)$ and the LLM backbone $\mathcal{LLM}(\cdot)$ are both finetuned to perform various visual tasks based on visual tokens containing different types of information. In this stage, only the vision encoder is kept frozen, and the multimodal projection layer and the LLM are trained in an end-to-end manner.
\end{itemize}

\subsection{Token reduction}

\paragraph{Mask Token Pruning}
In total, we generate 101 (100 object masks plus one background mask) segmentation masks to obtain their respective object tokens. However, many of these masks exhibit significant redundancy, capturing overlapping or highly similar regions within the image. To optimize computational efficiency and reduce the number of input tokens for the LLM, we can apply a mask filtering strategy based on Intersection over Union (IoU), by first removing all masks with a pixelwise IoU $\geq0.5$ and, second, ordering the remaining masks by their objectness confidence and further pruning the least confident masks.

We can apply the mask token pruning at training as well as at test time. Specifically, at test time, we prune the mask object tokens as described independent of how the model was trained and dynamically adjust the number of masks during inference. We provide a detailed analysis of this filtering strategy in Evaluation section \ref{sec:evaluation}.

\paragraph{Patch Token Pruning and pooling}
While the model is trained using patch features represented by $6\times6$ tokens (a total of 36 tokens), we can also further reduce the number of patch tokens at test time. This can be achieved by pruning, thus selecting a subset of patch tokens from the 36 available tokens or via additional average pooling over a fixed window to aggregate information. During the evaluation, we employ both strategies to reduce the number of tokens below 36 and to approximate the number of tokens for common benchmarks, showing the versatility and robustness of the pretrained model. 

\section{Evaluation}
\label{sec:evaluation}

\subsection{Datasets}
We conduct extensive experiments on eight image-based multimodal benchmarks on common vision language benchmarks, namely VQAv2 \cite{vqav2}, GQA \cite{gqa}, POPE \cite{pope}, MME \cite{mme}, MMBench \cite{mmbench}, ScienceQA-IMG \cite{scienceqa}, VizWiz \cite{vizwiz}, and MM-Vet \cite{mmvet}: \\
\textbf{VQAv2 \cite{vqav2}} evaluates the visual perception capabilities of VLMs through open-ended questions. It consists of 265,016 images, covering a variety of real-world scenes and objects. For each question, there are 10 ground truth answers provided by human annotators. \\
\textbf{GQA \cite{gqa}} is composed of three parts: scene graphs, questions, and images. The questions are designed to test the understanding of visual scenes and the ability to reason about different aspects of an image. \\
\textbf{POPE \cite{pope}} evaluates object hallucination in models using binary questions on object presence in images. \\
\textbf{MME \cite{mme}} evaluates the models performance across 14 subtasks targeting perceptual and cognitive abilities. \\
\textbf{MMBench \cite{mmbench}} evaluates models through three hierarchical levels of abilities: L-1 with two core abilities (perception and reasoning), L-2 with six sub-abilities, and L-3 with 20 specific dimensions.\\
\textbf{VizWiz \cite{vizwiz}} comprises of visual questions created by blind individuals, each capturing a photo using a mobile phone and recording a spoken question about it. The images are often of lower quality and some visual questions cannot be answered due to the nature of the content.\\
\textbf{ScienceQA \cite{scienceqa}} evaluates multimodal understanding, multi-step reasoning and interpretability. It has questions from 26 topics, 127 categories and 379 skills. \\
\textbf{MM-Vet \cite{mmvet}} focuses on the compilation of different vision-language capabilities, including recognition, OCR, knowledge, language generation, spatial awareness, and mathematics. It includes a total of 218 image-question pairs.

\subsection{Setup}
\paragraph{Segmentation Masks}
We first use the DDeTR framework \cite{deformable_detr_2020} to detect objects and their corresponding bounding boxes within the image, keeping the 100 bounding boxes along with their confidence scores per image. Then, we apply SAM~\cite{sam_iccv_2023} to generated segmentation masks based on these bounding boxes. We finally compute an additional background mask by first combining all the masks (pixelwise\_or), and obtaining the pixels not covered by (pixelwise\_not) based on the original image.

\paragraph{Training Details}
The training follows the two-stage pipeline of LLaVA, with vision-language pertaining followed by instruction fine-tuning. Namely, we first pretrain the projection layer on 90\% of LLaVA pre-training data (492K image-text pairs) for one epoch with a learning rate of 1e-3 and a global batch size of 256. Second, we finetune both, the projector and the language model, on LLaVA-Instruct (we use all the image-instruct pairs - 660K which do not have corrupted images) for one epoch with learning rate of 2e-5 and global batch size of 128. If not indicated otherwise, we use a single model pretrained and finetuned on an input of $36$ patch tokens, $1$ CLS token, and $101$ mask tokens for all experiments and only vary the number of tokens at test time without retraining or adapting the model.   


\begin{table*}[!ht]
    \centering
    \resizebox{\textwidth}{!}{%
    \begin{tabular}{@{}lcc|cccccccr@{}}
    \toprule
    Methods & RR  &  \# Vis. tokens & VQAv2 & GQA & POPE & MME & MMBench & SciQA & Vizwiz & MM-Vet\\ 
    \midrule
     LLaVA-1.5-7B \cite{visual_intruction_tuning_2023} & 0\% & 576 & 78.5 & 62.0 & 85.9 & 1510.7 & 64.3 & 66.8 & 50.0 & 30.5 \\ \midrule\midrule
     LLaVA-1.5-7B$\dagger$ \cite{visual_intruction_tuning_2023} & 90\%  & 58 & - & 54.2 & 74.6 & 1246.8 & 53.4 & 67.1 & - & 27.0\\ 
     FitPrune \cite{fitprune_2024} & 90\%  & 58 & 62.7 & 49.9 & 53.8 & 1147.4 &  56.2 & 68.2 & 50.8 & 21.8\\
     SparseVLM \cite{sparsevlm_2024} & 90\% & 58 & 62.9 & 48.8 & 65.8 & 1030.6 & 49.0 & 67.2 & 49.3 & 18.6\\ 
     FasterVLM \cite{faster_vlm_2024} & 90\%  & 58 & {71.9} & 54.9 & {75.8} & {1348.6} & 60.5 & {68.9} & {53.0} & {30.1}\\ \midrule
     MQT \cite{mqt_nips_2024} & 88\%  & 64 & 75.3 & 60.0 & {83.6} & \textbf{1464.3} & \textbf{63.5} & 67.0 & \underline{51.5} & \textbf{28.9} \\
     Voco-LLaMa \cite{vocollama_cvpr_2025}  & 88\%  & 64 & 75.4 & \underline{60.4} & - & - & 60.5 & - & - & - \\ 
     \textbf{Mask-LLaVA (ours)}  & 90\% & 57 & {74.8} & \textbf{60.6} & \underline{83.7} & \underline{1415.0} & \underline{63.1} & \underline{68.8} & \textbf{51.8} & 24.9 \\\midrule\midrule
     LLaVA-1.5-7B$\dagger$ \cite{visual_intruction_tuning_2023}& 95\% & 29 & - & 51.0 & 65.9 & 1141.1 & 45.7  & 67.1 & - & 23.5 \\
     FitPrune \cite{fitprune_2024} & 95\% & 29 & 52.3 & 43.6 &  31.1 & 855.2 & 39.6 & 68.3 & 48.6 & 18.0\\
     FasterVLM \cite{faster_vlm_2024} & 95\% & 29 & {66.7} & {51.5} & {67.2} & {1254.8} & {58.5} & {69.5} & {52.6} &  {27.5}\\  \midrule
     MQT \cite{mqt_nips_2024} & 88\%  & 36 & 73.7 & 58.8 & 81.9 & \underline{1416.3} & {63.4} & 66.8 & 51.0 & \textbf{27.8} \\
     M3 \cite{m3_iclr_2025} & 75\%  & 36 & 76.9 & \underline{60.3} & \textbf{85.5} & \textbf{1417.2} & \textbf{64.8} & 68.2 & \textbf{52.8} & 25.4 \\
     Voco-LLaMa \cite{vocollama_cvpr_2025} & 88\%  & 32 & 75.3 & 60.2 & - & - & 59.4 & - & - & - \\ 
     \textbf{Mask-LLaVA (ours)} & 97\% & 42 & 74.8 & \textbf{60.5} & {83.8} & {1402.7} & 63.2 & \underline{68.8} & \underline{52.1}  & \underline{25.7}\\ 
     \midrule\midrule
     MQT \cite{mqt_nips_2024} & 88\%  & 16 & 71.1 & 57.6 &  80.8 & \textbf{1408.5} & 61.9 & {67.5} & 49.8 & \textbf{25.3} \\
     M3 \cite{m3_iclr_2025} & 75\%  & 9 & 74.2 & 58.0 & \textbf{83.4} & 1374.8 & \textbf{63.1} & 68.0 & \underline{51.9} & 21.7 \\
     Voco-LLaMa \cite{vocollama_cvpr_2025} & 88\%  & 16 & \textbf{75.4} & \textbf{59.4} & - & - & 58.6 & - & - & - \\ 
     \textbf{Mask-LLaVA (ours)} & 97\% & 15 & \underline{71.5} & \underline{58.5} & \underline{82.1} & \underline{1395.8} & \underline{62.1} & \textbf{68.4} & \textbf{52.8} & \underline{21.9}\\ 
     
    \bottomrule
    \end{tabular}
    }
    \caption{\textbf{Performance comparison of different methods on 8 benchmarks at reduction ratios ranging from $75\%$ to $\ge97\%$}. For each reduction ratio (RR), the best performance is shown in \textbf{bold} and the second best performance is shown in \underline{underline}. The proposed Mask-LLaVA model is especially competitive for reduction ratios $\geq90\%$, outperforming other methods, especially on $42$ and $29$ tokens.  \# Vis. tokens indicates the number of vision tokens fed to LLM backbone; $\dagger$: patch tokens randomly dropped at test time; $\ddagger$: results from \cite{llava_mini_2025} without modality fusion. }
    
    \label{tab:main_vqa_comparison}
\end{table*}

\subsection{Comparison to State-of-the-Art}

We first compare the resulting model to other efficient VLM approaches. Table~\ref{tab:main_vqa_comparison} shows the zero-shot performance of the considered methods for LLaVA-1.5-7b model as the visual token reduction ratio increases from 75\% to $\ge95\%$. 
For Mask-LLaVA, we leverage the backbone trained on 138 tokens $(36+1+101)$ and first gradually reduce the number of mask tokens used at test time to $20$ resp. $5$ leading to $57$ resp $42$ tokens overall. 
Second, for the last two setups, we further reduce the number of patch tokens, first via pruning them to 23, leading to 29 tokens overall $(23+1+5)$, as well as via an additional $2\times2$ spatial average pooling resulting in $15$ tokens composed as $(9+1+5)$. 

To directly compare with the original LLaVA-1.5-7B model~\cite{Improved_baselines_with_visual_instruction_tuning_cvpr_2024}, we reduce the visual token footprint by randomly dropping tokens at test time. 
Out of the other models, FastV~\cite{fastv_eccv_2024}, FitPrune~\cite{fitprune_2024}, SparseVLM~\cite{sparsevlm_2024}, and FasterVLM are also pruned at the test time, while LLaVA-Mini~\cite{llava_mini_2025} rely on a token compression module individually trained for each different token reduction setting. 
Note that LLaVA-Mini~\cite{llava_mini_2025} also proposes a modality fusion module in addition to token compression. As the modality fusion can be considered an independent architectural component which can be considered complementary to the problem of vision token reduction, we consider here only results without modality fusion for direct comparability. 

Overall it shows that the proposed Mask-LLaVA model is mainly able to achieve state-of-the-art performance for a reduction ratio of $\ge90\%$, e.g. reaching best performance on four out of eight benchmarks, namely GQA, POPE, MME, and MMBench and achieving second best results on two more, ScienceQA and MM-Vet for $57$ tokens,  and even outperfroming on five out of eight benchmarks, namely VQAv2, GQA, POPE, MME, and MMBench and achieving second best results on two more, Vizwiz and MM-Vet for $29$ tokens. 

Mask-LLaVA further achieves highly competitive results when comparing the currently best performance on $58$ tokens with the performance of Mask-LLaVA with only $42$ tokens where even with this reduced token size, it still outperforms models with more tokens on five datasets. 

Regarding absolute performance, it shows that especially on POPE and MME, Mask-LLaVA is able to increase the state-of-the-art by a significant margin. 

Regarding the overall behavior of the model, it becomes clear that the higher performance can be mainly attributed to the fact that the model drops much less from on reduction level to the next compared to other models, with performance decreasing often only around $1\%$.   
This can to a certain extend be attributed to the local patch tokens, but even when starting to reduce those as well, for the cases of $29$ and $15$ tokens, the overall performance still holds, challenging even trained compression methods like LLaVA-Mini.

\subsection{Ablation}
\paragraph{Types of masks.} Firstly, we consider the type of masks we feed as an input to the Mask token computation module. The mask token computation module is independent of the quality of the masks we feed. To validate this with experiments, we consider three types of masks: (a) Masks generated by SAM using the bounding boxes from DeTR, (b) Masks generated by using just the bounding boxes generated by DeTR, and (c) Masks generated by converting an image into tiles with different sizes (example: 2x2, 3x3, 4x4, and 5x5 tiles). From Table. \ref{tab:types_of_masks}, it can be seen that, the performance is maintained across both POPE \cite{pope} and MME \cite{mme} datasets, showing that the method is robust and can work with any type of mask.

\begin{table}[!ht]
    \centering
    \resizebox{0.48\textwidth}{!}{%
    \begin{tabular}{@{}ll|ccc@{}}
    \toprule
    Types of tokens & \#vis tok & POPE & MME  \\
    \midrule
    Detr + SAM  &  150  & 86.33 & 1391.01 \\
      & 42 &  85.33 & 1407.37 \\ 
      & 15 &  82.94 & 1357.75 \\ 
     \midrule
    Detr masks  & 150 &  - & 1395.39  \\
      & 42  &  - & 1399.02 \\ 
     & 15 &  - & 1353.56  \\ 
     \midrule

    Tiled masks  & 154  & 86.26 & 1401.62\\
      & 41  & 85.39 & 1409.39 \\ 
      & 14  & 82.96 & 1360.03 \\ 
      
      \bottomrule
    \end{tabular}
         }
    \caption{\textbf{Ablation Study: Different types of masks.} In this table, we evaluate the performance of the method (epoch1) based on the type of masks we use to compute maskinversion tokens across POPE \cite{pope}, and MME \cite{mme} datasets.}
    \label{tab:types_of_masks}
    \vspace{-2.5mm}
\end{table}

\paragraph{Impact of Token Composition}
Second, we consider the token composition. To this end, we leverage the original backbone trained on 138 tokens and drop the respective token types at test time, namely testing the backbone with only patch tokens, patch and CLS tokens, patch and mask tokens, as well as with a combination of all in Table \ref{tab:token_composition_ablation}. 

Here it shows that even the original backbone trained with $6x6$ average pooled tokens already provides a strong baseline. It further shows that adding scaled CLS respectively mask tokens increase this baseline in different ways. While e.g. the CLS token provides an important signal for the MME and ScienceQA benchmark, the mask tokens rather support the performance on POPE and both together increase the performance for POPE and ScienceQA indicating the complementary nature of those two information. Overall it shows that the triplet of the respective token types provides better results than the patch tokens alone.

\begin{table}[!ht]
    \centering
    \resizebox{0.48\textwidth}{!}
    {%
    \begin{tabular}{@{}ll|cc@{}}
    \toprule
    Type & \#vis tok  & POPE & MME \\
    \midrule
    {Only Patch Tokens}    & 144   & 85.14 & 1444.77 \\ 
    {Only Patch Tokens}    & 36   & 83.47 & 1411.74 \\
    {Only Patch Tokens}    & 16   & 82.79 & 1385.27 \\
    {Only Patch Tokens}    & 9   & 81.29 & 1380.10 \\ \midrule
    {Patch + [CLS]}        & 144+1 & - & 1451.97 \\ 
    {Patch + [CLS]}        & 36+1 & - & 1407.28 \\
    {Patch + [CLS]}        & 16+1 & - & 1388.18 \\
    {Patch + [CLS]}        & 9+1 & - & 1408.79 \\ \midrule
    {Patch + [CLS] + Mask tokens}  & 144+1+5 & 86.33 & 1391.01 \\
    {Patch + [CLS] + Mask tokens}  & 36+1+5 & 85.33 & 1407.37 \\
    {Patch + [CLS] + Mask tokens}  & 16+1+5 & 84.42 & 1408.12 \\
    {Patch + [CLS] + Mask tokens}  & 9+1+5 & 82.94 & 1357.75 \\
    \bottomrule
    \end{tabular}
    }
    \caption{\textbf{Ablation Study: Token Composition.} In this table, we present the performance of Mask-LLaVA when pretrained and fine-tuning on 138 tokens. Specifically, 36 patch tokens, the [CLS] token, and 101 mask object tokens. The model is then tested with individual feature types and their combinations. }
    \label{tab:token_composition_ablation}
    \vspace{-2.5mm}
\end{table}

\paragraph{Visual in-context prompting - ViP-Bench - TBD}

\begin{figure*}[t]
  \centering
   \includegraphics[width=1.0\linewidth]{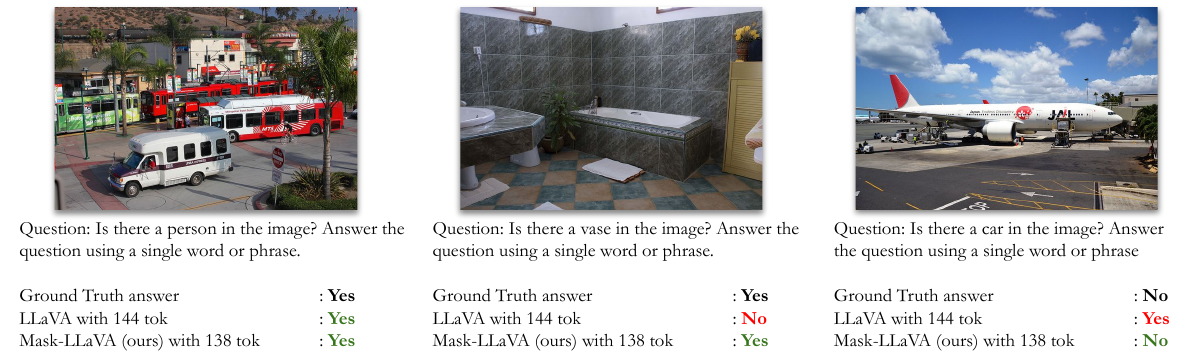}

   \caption{\textbf{Qualitative Results for POPE \cite{pope} dataset.}}
   \label{fig:pope_only_v2}
\end{figure*}

\begin{figure*}[t]
  \centering
   \includegraphics[width=1.0\linewidth]{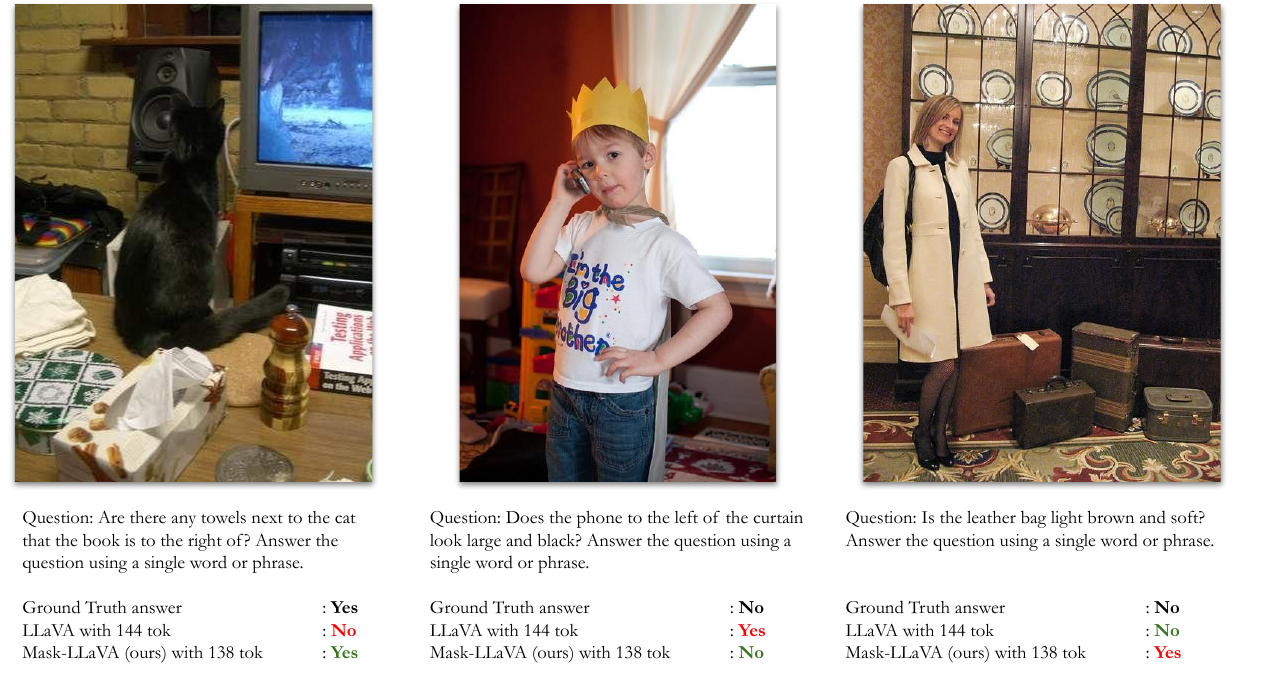}

   \caption{\textbf{Qualitative Results for GQA \cite{gqa} dataset.}}
   \label{fig:qual2_gqa}
   \vspace{-2mm}
\end{figure*}

\subsection{Qualitative examples}

We analyze qualitative results from the POPE and GQA benchmarks, comparing answers produced by Mask-LLaVA (trained with 138 tokens and tested with varying token counts) against the original LLaVA backbone (using 144 tokens at test time).
Note that the examples present a random selection and were not specifically picked. For POPE, shown in Figure~\ref{fig:pope_only_v2}, we can observe that Mask-LLaVA performs well with respect to its robustness to hallucinations e.g. being able to spot correctly if an object is present in an image or not. Further especially on the POPE dataset, we observe that the pooling of patch tokens does not result in a loss of performance with respect to smaller fine-grained objects, which might be compensated by the respect mask embeddings. 
For GQA, shown in Figure~\ref{fig:qual2_gqa}, we can see that Mask-LLaVA further performs well with respect to the presence of the object and their attributes such as color showing, also potential to connect those attributes with location information to capture the composition of a scene.
For the third example in this dataset ("Is the leather bag light brown and soft?"), Mask-LLaVA interestingly fails to provide the correct answer. This limitation may stem from CLIP embeddings (used for both the CLS token and as the basis for mask tokens) functioning additively, thus evaluating whether either condition is fulfilled rather than interpreting the question as a logical AND conjunction. More qualitative examples can be found in the appendix.

\section{Conclusion}
\label{conclusion}
We presented Mask-LLava, an efficient VLM framework that leverages visual token of different representations granularity, namely, CLS tokens, pooled patch tokens, as well as mask-based object tokens. Our evaluation shows that a model trained with a scaled, concatenated input of those tokens is highly robust to simple token reduction techniques like pruning and pooling at test time opening up the possibility of state-of-the-art VLM performance on devices with less computational capabilities as well as for more complex scenarios with multitudes of images and text.  


\section*{Acknowledgments}
Soumya Jahagirdar is funded by the European Research Council (ERC) under the Starting Grant \emph{GraViLa} (101117556). Walid Bousselham is supported by the German Federal Ministry of Education and Research (BMBF) through the project \emph{STCL} (01IS22067) and by the T\"ubingen AI Center. The authors gratefully acknowledge the Gauss Centre for Supercomputing e.V.\ (www.gauss-centre.eu) \cite{juwels} for funding this work by providing computing time on the GCS Supercomputer JUPITER---JUWELS at the J\"ulich Supercomputing Centre (JSC).
This work also acknowledges support from the project \emph{ELLIOT-FM: Open Multi-Modal Foundation Models with Strong Generalization and Reasoning}.


\bibliography{tmlr}
\bibliographystyle{tmlr}

\appendix
\section{Appendix}

\section{Token Composition in Mask-LLaVA}
\begin{figure*}[htbp]
    \centering
    \begin{minipage}[b]{0.75\textwidth}  
        \centering
        \includegraphics[width=\linewidth]{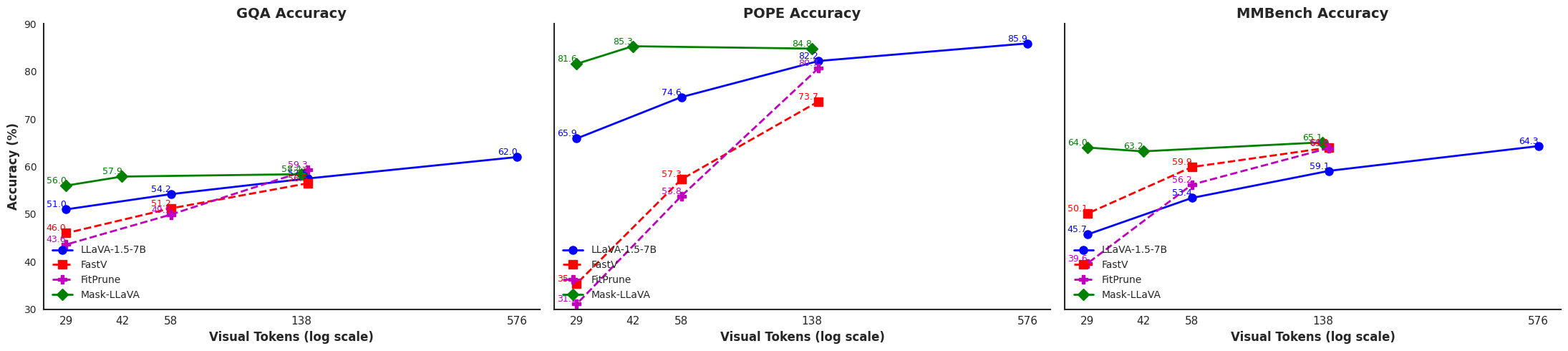}
        \label{fig:gqa_pope_mmbench}
    \end{minipage}%
    \hfill
    \begin{minipage}[b]{0.25\textwidth}  
        \centering
        \includegraphics[width=\linewidth]{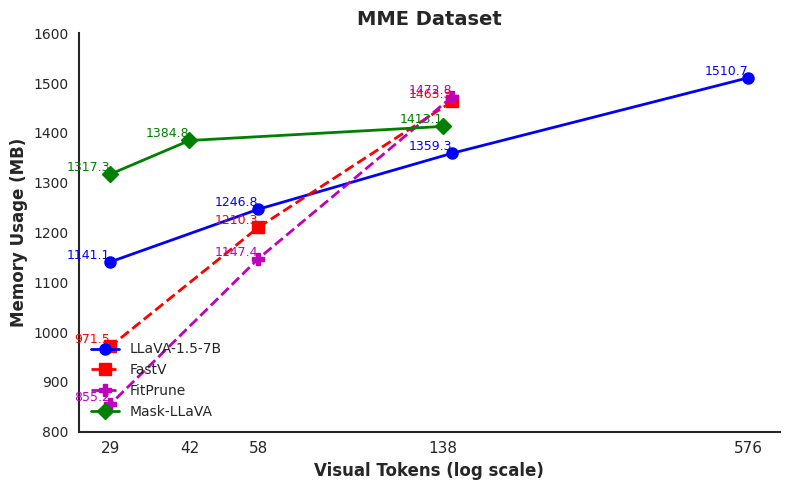}
        \label{fig:mme}
    \end{minipage}
    \caption{\textbf{Performance Comparison of LLaVA-1.5-7b \cite{visual_intruction_tuning_2023}, FastV \cite{fitprune_2024}, FitPrune \cite{fitprune_2024}, and Mask-LLaVA (ours).} In this figure, we compare the performance of LLaVA at different token reduction rates during inference, alongside the FastV and FitPrune methods, with Mask-LLaVA. The results show that Mask-LLaVA exhibits strong performance even with a small number of tokens, and its performance trend consistently improves as the number of tokens increases. This comparison is present across four different datasets.}
    \label{fig:combined}
\end{figure*}

Figure. \ref{fig:combined} presents a performance comparison of LLaVA under different token reduction rates during inference alongside the FastV \cite{fastv_eccv_2024}, FitPrune \cite{fitprune_2024}, and Mask-LLaVA. The results demonstrate that our approach remains highly competitive even when using a minimal number of tokens. Additionally, other methods exhibit a sharper performance decline with fewer tokens, highlighting the importance of token composition in Mask-LLaVA.


\section{Additional Experiments/Details}

\subsection{Mask-LLaVA with Vicuna-13B}

\begin{table*}[!ht]
    \centering
    \resizebox{\textwidth}{!}{%
    \begin{tabular}{@{}lcc|ccccccr@{}}
    \toprule
    Methods & Reduction Ratio  &  \# Vis. tokens & GQA & POPE & MME & MMBench & SciQA & Vizwiz & MM-Vet\\ 
    \midrule
    LLaVA-1.5-13B & 0 & 576  & 63.3 & 85.9 & 1531.3  & 67.7 & 71.6 & 53.6 & 36.1 \\ \midrule
    Mask-LLaVA-13B (ours) & 75\% & 138  & 60.2 & 85.5 & 1439.6 & 64.9 & 70.8 & 49.6 & 31.1\\
    Mask-LLaVA-13B (ours) & 90\% & 57  & 59.6 & 85.8 & 1434.9 & 64.6 & 70.6 & 50.1 & 29.0\\
    Mask-LLaVA-13B (ours) & 95\% & 29  & 58.4 & 85.6 & 1422.6 & 64.0 & 70.6 & 51.8 & 30.0\\

    \bottomrule
    \end{tabular}
    }
    \caption{\textbf{Performance comparison of different methods on 8 benchmarks at reduction ratios ranging from $75\%$ to $\ge97\%$}. Here, the base LLM is Vicuna-13B \cite{vicuna_2023}.}
    \label{tab:many_vqa_13b}
\end{table*}

Table. \ref{tab:many_vqa_13b} presents the performance of Mask-LLaVA with varying token reduction ratios across multiple benchmarks, including GQA \cite{gqa}, POPE \cite{pope}, MME \cite{mme}, MMBench \cite{mmbench}, SciQA \cite{scienceqa}, Vizwiz \cite{vizwiz}, and MM-Vet\cite{mmvet}. Our model is trained using 36 patch tokens, [CLS], and mask object tokens, where the latter are scaled based on the mean and standard deviation of the patch norms corresponding to the image. Mask-LLaVA-13B model follows the same hyperparameters for pretraining and fine-tuning as the original Mask-LLaVA-7B. During pretraining, only the multimodal projector is trained, while in fine-tuning, both the Vicuna-13B and the multimodal projector are trained. It can be seen from the Table. \ref{tab:many_vqa_13b} that, even with as few as 29 tokens, our method maintains strong performance across most datasets.

\begin{table*}[!ht]
    \centering
    \resizebox{\textwidth}{!}{%
    \begin{tabular}{@{}lc|ccccccccccr@{}}
    \toprule
    Methods & \# Vis. tok & Existence  & Count & Position & Color & Posters & Celebrity & Scene  & Landmark & Artwork & OCR\\ 
    \midrule
    LLaVA-1.5 & 576 & 190.0 & 143.3 & 123.3 & 150.0 & 169.0  & 113.2 & 160.2 & 171.5 & 119.2 & 95.0 \\  \midrule
    LLaVA-1.5 & 144 & 185.0 & 130.0 & 123.3 & 130.0 & 141.1  & 120.5 & 166.2 & 147.5 & 108.0 & 107.5 \\ 
    Mask-LLaVA & 138 & 195.0 & 140.0 & 118.3 & 145.0 & 155.4  & 102.3 & 157.0 & 158.0 & 119.5 & 122.5 \\ \midrule
    LLaVA-1.5 & 58 & 170.0 & 120.0 & 81.6 & 118.3 & 124.8  & 98.2 & 169.7 & 131.5 & 100.0 & 132.5 \\ 
    Mask-LLaVA & 57 & 185.0 & 140.0 & 113.3 & 145.0 & 154.4  & 105.2 & 154.7 & 158.7 & 121.0 & 122.5 \\ \midrule
    LLaVA-1.5 & 29 & 145.0 & 98.3 & 93.3 & 108.3 & 127.2  & 87.6 & 163.0 & 119.2 & 89.0 & 110.0 \\ 
    Mask-LLaVA & 29 & 190.0 & 145.0 & 110.0 & 135.0 & 147.6  & 98.5 & 154.7 & 158.7 & 117.7 & 127.5 \\
    \bottomrule
    \end{tabular}
    }
    \caption{\textbf{MME Benchmark – Perception Evaluation.} This table presents the performance comparison of LLaVA and Mask-LLaVA on the perception category of the MME benchmark. Notably, Mask-LLaVA maintains strong performance even with fewer visual tokens, demonstrating its ability to retain essential information while reducing the number of visual tokens.}
    \label{tab:mme_perception}
\end{table*}

\begin{table*}[!ht]
    \centering
    \resizebox{0.75\textwidth}{!}{%
    \begin{tabular}{@{}lc|cccr@{}}
    \toprule
    Methods & \# Vis. tok & \makecell[l]{Commonsense \\ Reasoning}  & \makecell[l]{Numerical \\ Calculation} & \makecell[l]{Text \\ Translation} & \makecell[l]{Code \\ Reasoning}\\ 
    \midrule
    LLaVA-1.5 & 576 & 123.5 & 47.5 & 80.0 & 55.0 \\ \midrule
    LLaVA-1.5 & 144 & 106.4 & 57.5 & 92.5 & 45.0 \\ 
    Mask-LLaVA & 138 & 127.1 & 62.5 & 57.5 & 45.0 \\  \midrule
    LLaVA-1.5 & 58 & 106.4 & 72.5 & 65.0 & 67.5 \\  
    Mask-LLaVA & 57 & 132.1 & 55.0 & 62.5 & 40.0 \\  \midrule
    LLaVA-1.5 & 29 & 109.2 & 72.5 & 57.5 & 45.0 \\
    Mask-LLaVA & 29 & 125.0 & 55.0 & 50.0 & 50.0 \\
    \bottomrule
    \end{tabular}
    }
    \caption{\textbf{MME Benchmark – Cognition Evaluation.}This table compares the performance of LLaVA and Mask-LLaVA on the cognition category of the MME benchmark, where Mask-LLaVA continues to perform effectively with fewer visual tokens, showcasing its efficient handling of complex tasks.}
    \label{tab:mme_cognition}
\end{table*}

\subsection{MME Benchmark}

Table. \ref{tab:mme_perception} and Table. \ref{tab:mme_cognition} compares the performance of LLaVA with Mask-LLaVA using different numbers of input tokens for MME Benchmark \cite{mme}. The reported values correspond to multiple classes in the perception category of MME Benchmark \cite{mme}. 

\subsubsection{Perception Tasks}
Coarse-grained Recognition involves identifying common objects along with their count, color, and position. Images are sources from COCO, but instruction-answer pairs are manually created rather than derived from existing annotations. This ensure that even if MLLMs have encountered these images before they have not seen the specific instruction-answer pairs during training. The task evaluates the model's ability to interpret instructions and infer answers. In each perception subtask of existence, count, color, and position 30 images with 60 instruction-answer pairs are present. Fine-grained Recognition evaluates the knowledge capabilities of MLLM across various categories including movie posters, celebrities, scenes, landmarks, and artworks, with 147, 170, 200, 200 and 200 images respectively. Similar to coarse-grained recognition, all the instructions are manually generated. For OCR, 20 images are manually annotated with 40 designed instruction-answer pairs.

From Table. \ref{tab:mme_perception}, Mask-LLaVA outperforms LLaVA when tested with a reduced number of visual tokens, specifically with 144, 58, and 29 tokens. Interestingly, in some cases, Mask-LLaVA with just 29 tokens achieves better performance than with 57 tokens, suggesting that a smaller test of tokens is crucial for accurate predictions of certain categories. Similarly, from Table. \ref{tab:mme_cognition}, it can be seen that Mask-LLaVA outperforms LLaVA with different visual token inputs.

\begin{table}[!ht]
    \centering
    \resizebox{0.48\textwidth}{!}{%
    \begin{tabular}{@{}l|cc@{}}
    \toprule
    Category & LLaVA-1.5 & Mask-LLaVA  \\ 
    \midrule
    Action Recognition & 59.2 & 83.3  \\ 
    Attribute Comparison & 56.8 & 56.8 \\  
    Attribute Recognition & 56.7 & 77.0  \\  
    Celebrity Recognition & 49.4 & 81.8  \\ 
    Function Reasoning & 59.4 & 81.0 \\
    Future Prediction & 25.0 & 32.5\\
    Identity Reasoning & 80.0 & 91.1\\
    Image Emotion & 48.0 &  78.0 \\
    Image Quality & 20.7 & 43.3  \\
    Image Scene & 73.0 & 97.1  \\
    Image Style & 41.5 & 73.5  \\
    Image Topic & 72.2 & 83.3  \\
    Nature Relation & 27.0 & 47.9  \\
    Object Localization & 25.9 & 39.5  \\
    Ocr & 30.7 & 53.8  \\
    Physical Property Reasoning & 49.3 & 53.3  \\
    Physical Relation & 16.6 & 29.1 \\
    Social Relation & 41.8 & 81.3  \\
    Spatial Relationship & 20.0 & 15.5 \\
    Structuralized image-text Understanding & 24.3 & 28.2  \\

    \bottomrule
    \end{tabular}
    }
    \caption{\textbf{MMBench Benchmark.} This table presents the performance comparison of LLaVA and Mask-LLaVA on the MMBench benchmark, where Mask-LLaVA demonstrates superior performance even with only 29 tokens at test time, highlighting its efficiency in handling reduced token inputs.} 
    \label{tab:mmbench}
\end{table}

\begin{table}[!ht]
    \centering
    \resizebox{0.48\textwidth}{!}{%
    \begin{tabular}{@{}lc|ccr@{}}
    \toprule
    Scaling & \#vis tok & MM-Vet & POPE & MME \\
    \midrule
     9+1+5 (avg. pool 8x8) & 15 & 21.3 & 81.1 & 1317.3\\ 
     9+1+5 (random 9) & 15  & 22.2 & 67.8 & 1112.1\\ 
     9+1+5 (maxpool 9) & 15  & 21.2 & 67.4 & 1195.4\\ 
      \bottomrule
    \end{tabular}
         }
    \caption{\textbf{Ablation Study on Different Token Reduction Styles.} This table compares the performance of Mask-LLaVA using various token reduction strategies, highlighting their impact on model accuracy across multiple benchmarks.}
    \label{tab:tok_reduction}
\end{table}

\begin{table}[!ht]
    \centering
    \resizebox{0.48\textwidth}{!}{%
    \begin{tabular}{@{}lc|cr@{}}
    \toprule
    Methods & \# Vis. tok & Accuracy  & IMG-Accuracy \\ 
    \midrule
    LLaVA-1.5 & 144 & 70.2 & 69.5 \\ 
    Mask-LLaVA & 138 & 71.9 & 70.5 \\  \midrule
    LLaVA-1.5 & 58 & 69.0 & 67.1 \\  
    Mask-LLaVA & 57 & 71.4 & 69.5 \\  \midrule
    LLaVA-1.5 & 29 & 69.1 & 67.1 \\
    Mask-LLaVA & 29 & 71.0 & 68.5 \\
    \bottomrule
    \end{tabular}
    }
    \caption{\textbf{SciQA \cite{scienceqa} Benchmark.} This table compares the performance of LLaVA and Mask-LLaVA on the SciQA benchmark, where Mask-LLaVA consistently outperforms LLaVA across different numbers of tokens at test time.}
    \label{tab:sciqa_benchmark}
\end{table}

\begin{table*}[!ht]
    \centering
    \resizebox{\textwidth}{!}{%
    \begin{tabular}{@{}lc|ccccccr@{}}
    \toprule
    Methods & \# Vis. tok & Recognition  & OCR & Knowledge & Language generation & Spatial Reasoning & Math & Total \\ 
    \midrule
    LLaVA-1.5 & 576 & 35.8 & 23.2 & 16.7 & 22.0 & 26.4  & 11.5 & 31.0 \\  \midrule
    LLaVA-1.5 & 144 & 33.7 & 17.9 & 21.1 & 22.4 & 25.2  & 11.2 & 38.2 \\ 
    Mask-LLaVA & 138 & 36.6 & 18.9 & 22.0 & 23.2 & 27.5  & 11.2 & 30.0 \\ \midrule
    LLaVA-1.5 & 58 & 32.6 & 14.6 & 20.4 & 20.5 & 26.3  & 11.2 & 27.0 \\ 
    Mask-LLaVA & 57 & 34.7 & 16.1 & 18.9 & 19.7 & 26  & 7.7 & 27.8 \\ \midrule
    LLaVA-1.5 & 29 & 28.7 & 15.0 & 17.6 & 18.1 & 24.5  & 3.8 & 23.5 \\ 
    Mask-LLaVA & 29 & 32.1 & 13.3 & 19.8 & 21.9 & 22.8  & 7.7 & 25.2 \\
    \bottomrule
    \end{tabular}
    }
    \caption{\textbf{MM-Vet Benchmark.}This table presents the performance comparison of LLaVA and Mask-LLaVA on the MM-Vet \cite{mmvet} benchmark, where Mask-LLaVA performs better across different numbers of tokens at test time. The scores for various categories, including recognition, OCR, knowledge, language generation, spatial reasoning, and math, are also provided, highlighting the comprehensive capabilities of Mask-LLaVA with fewer visual tokens.}
    \label{tab:mmvet}
\end{table*}

\subsection{MMBench benchmark}

MMBench evaluates the models with different hierarchical levels of abilities. L-1 with two core abilities (perception and reasoning), and L-2 with six sub-abilities, and L-3 with 20 specific dimensions. In Table. \ref{tab:mmbench}, we show the performance of LLaVA-1.5 with 29 tokens at the time of testing, and Mask-LLaVA with 29 tokens at the time of testing. It can be seen that on various categories, Mask-LLaVA outperforms LLaVA-1.5 by a good margin.

\subsection{MM-Vet Benchmark}
This benchmark, defines six core vision-and-language capabilities which are recognition, OCR, knowledge, language generation, spatial awareness, and math. Further, these capabilities are integrated to address a range of complex multimodal tasks with 16 specific integrations of these capabilities. In Table. \ref{tab:mmvet}, we show the performance comparison of LLaVA-1.5 and Mask-LLaVA with different token reduction ratios. It can be seen that, with the drop in the number of visual tokens, LLaVA suffers immensely, whereas Mas-LLaVA still performs decently with 29 visual tokens.

\subsection{SciQA Benchmark}

This benchmark spans domains like natural, language, and social science. The questions require multimodal understanding, multimodal understanding, multi-step reasoning and interpretability. In Table. \ref{tab:sciqa_benchmark}, we show the performance of LLaVA-1.5 and Mask-LLaVA at different token reduction rates. Similar to other datasets, Mask-LLaVA still maintains the performance even with the lower number of visual tokens.

\subsection{Token Reduction Strategy}

To evaluate the performance of Mask-LLaVA at the time of testing, as mentioned in the main section, we reduce the number of tokens at the time of testing. We can either drop local patch features, or object features or both. In Table. \ref{tab:tok_reduction}, we show the performance of Mask-LLaVA with three different types of token reduction styles. Initially, we have 576 total patch tokens, and these token reduction styles are used to further reduce the number of patch tokens. For row 1, avg. pool 8x8, we apply a kernel size of 8x8 over the 24x24 2D grid obtained from 576 patch tokens, which results in 9 pooled patch tokens. For row 2, we first apply average pooling using 4x4 over 24x24 2D grid, and then we randomly sample 9 tokens. For row 3, we apply max pooling over 36 patch tokens with a kernel size of 2x2. It can be seen that average pooling with 8x8 works best for two out of three datasets.

\subsection{Token Scaling}
As explained in the main paper, we have different types of scaling for the mask object representations and global representation [CLS] based on local patch features. We first normalize the mask object representations and [CLS] with their own mean and standard deviation and then proceed with scaling both based on mean and standard deviation of the local patch features.

\section{Qualitative Results}

\begin{figure*}[t]
  \centering
   \includegraphics[width=1.0\linewidth]{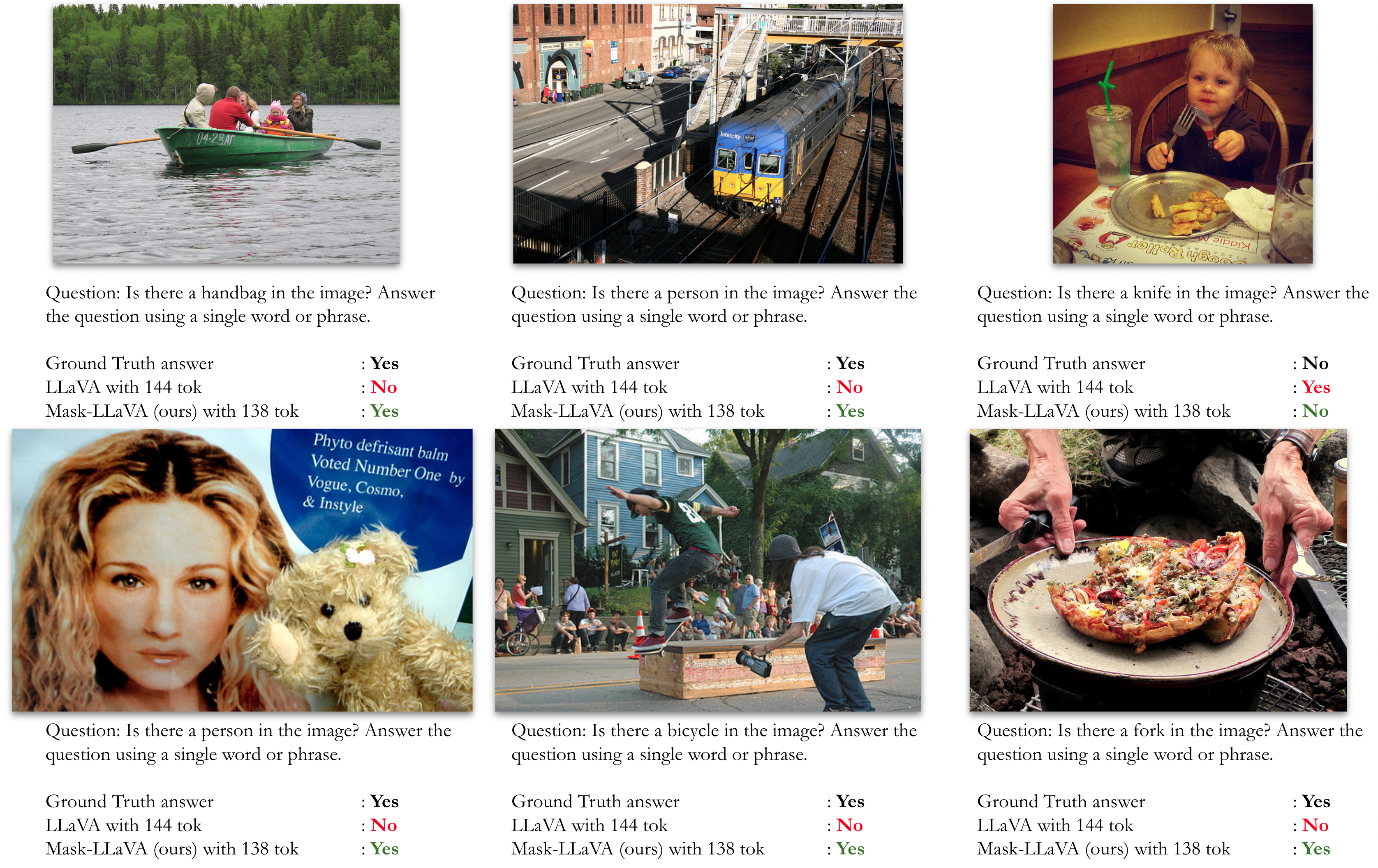}

   \caption{\textbf{POPE Qualitative Results.} Answers in green are correct generated response and answers in red are incorrect generated response by respective methods.}
   \label{fig:pope_qual1}
\end{figure*}

\begin{figure*}[t]
  \centering
   \includegraphics[width=1.0\linewidth]{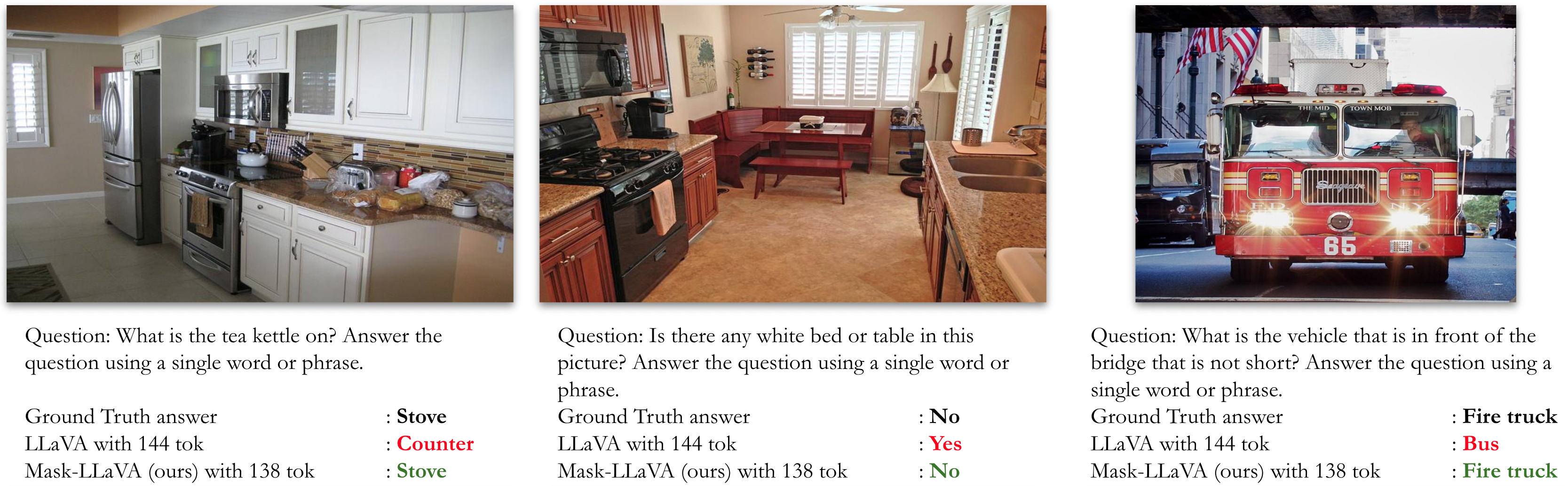}

   \caption{\textbf{GQA Qualitative Results.} Answers in green are correct generated response and answers in red are incorrect generated response by respective methods.}
   \label{fig:gqa_qualitative}
\end{figure*}

GQA Qualitative Results. Figure \ref{fig:gqa_qualitative} showcases qualitative comparisons between LLaVA and Mask-LLaVA on the GQA dataset. Correctly generated responses are highlighted in green, while incorrect ones are marked in red. As seen in the examples, Mask-LLaVA demonstrates a stronger ability to generate accurate answers, particularly in complex reasoning scenarios. Compared to LLaVA, Mask-LLaVA better captures fine-grained details and produces more precise responses, highlighting its improved understanding of visual and textual context.

POPE Qualitative Results. Figure \ref{fig:pope_qual1} presents a qualitative comparison of responses generated by LLaVA and Mask-LLaVA on the POPE benchmark. Correct responses are highlighted in green, while incorrect ones are marked in red. The results illustrate that Mask-LLaVA consistently provides more accurate answers, particularly in scenarios requiring precise perception and reasoning. Compared to LLaVA, Mask-LLaVA demonstrates a stronger ability to understand and interpret visual content, leading to improved performance on this benchmark.



\end{document}

%% file: math_commands.tex
\usepackage{amsmath,amsfonts,bm}

\def\eqref#1{equation~\ref{#1}}

\def\1{\bm{1}}

\DeclareMathAlphabet{\mathsfit}{\encodingdefault}{\sfdefault}{m}{sl}
\SetMathAlphabet{\mathsfit}{bold}{\encodingdefault}{\sfdefault}{bx}{n}

